\documentclass[11pt]{article}
\usepackage[final]{acl}
\usepackage{times}
\usepackage{latexsym}
\usepackage[T1]{fontenc}
\usepackage[utf8]{inputenc}
\usepackage{microtype}
\usepackage{inconsolata}
\usepackage{graphicx}
\usepackage{subcaption}
\usepackage{booktabs}
\usepackage{xcolor}
\usepackage{amsmath}
\usepackage{amssymb}
\usepackage{tikz}
\usetikzlibrary{arrows.meta,positioning,fit,calc}

\usepackage[normalem]{ulem}   
\newcommand{\sys}{AgentGUI}
\newcommand{\code}[1]{\texttt{\small #1}}
\newcommand{\panel}[1]{{\normalfont\bfseries\fontsize{10}{12}\selectfont #1.}}

\title{\sys{}: An Interface for Observing and Steering\\Long-Running AI Agents}
\author{
  Xuan Zhao$^{1}$ \quad Jiwoong Sohn$^{1,2}$ \quad
  Qinyue Zheng$^{1,2}$ \quad Michael Moor$^{1,2}$ \\[0.6em]
  $^{1}$ETH Zürich \quad
  $^{2}$ETH AI Center \\
}

\begin{document}
\maketitle

\begin{abstract}
AI agents are increasingly adept at tackling complex, long-running tasks.
With the rapid surge of autonomous capabilities, human oversight is systematically lagging behind due to limited human-centered interfacing. Aiming to address this, we introduce \sys{}, a user-friendly, locally hosted GUI for seamlessly observing and steering AI agents amid multiple concurrent, long-running sessions. \sys{} features 1) rich agent trajectory visualizations, 2) effective manual and automated steering, and 3) integration with and coordination between open-source and frontier agent frameworks. A controlled user study demonstrates statistically significant reduction in the time it takes to identify key elements from agent traces ($38\%$ faster, $p = 0.023$). In a preliminary experiment, \sys{}'s automated drift prevention feature raises the task completion rate of small local agents by as high as $34$\,pp across a 0.8B--9B model ladder ($N{=}50$ runs per model). \sys{} is publicly available through its project website\footnote{\url{https://agent-gui-project.github.io/}} and open-source repository\footnote{\url{https://github.com/eth-medical-ai-lab/agent-gui}}, along with a demo video\footnote{\url{https://youtube.com/watch?v=GSDyxN1gTF0}}.
\end{abstract}

\begin{figure*}[t]
  \centering
  \includegraphics[width=\textwidth]{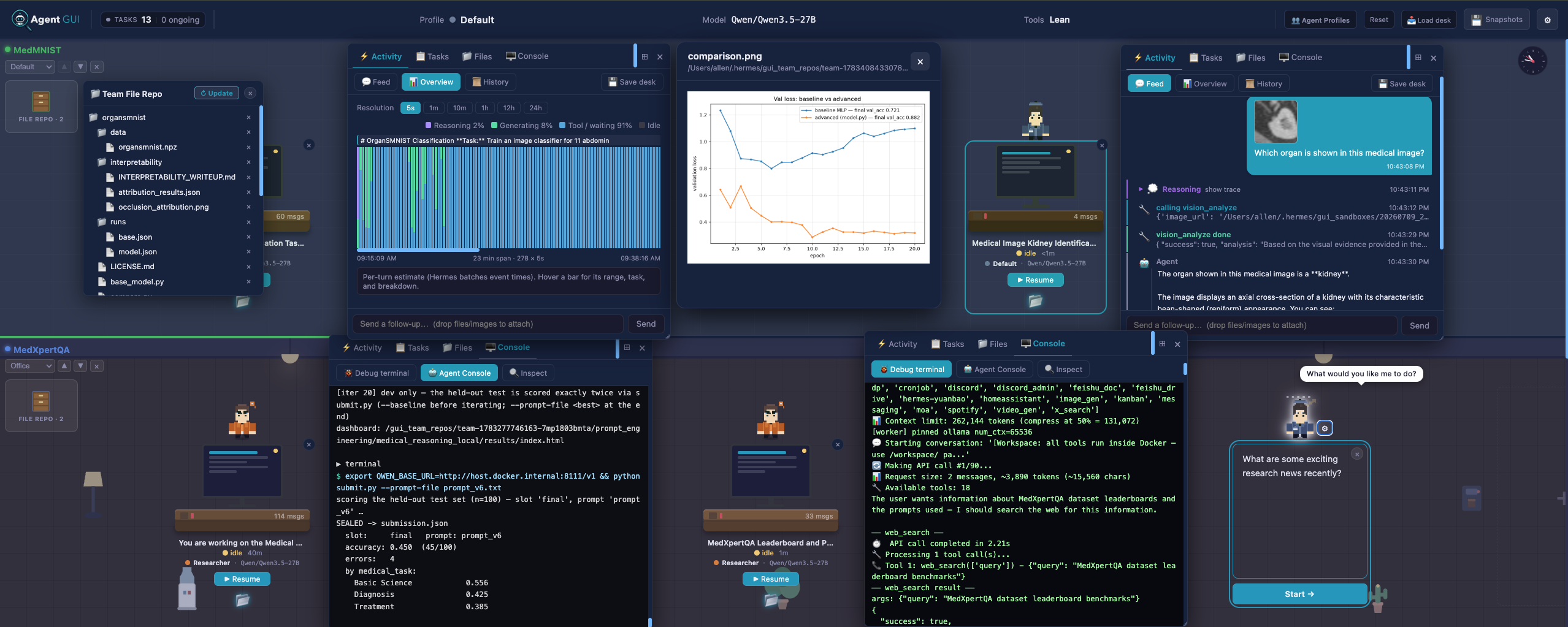}
  \caption{The \sys{} dashboard. The dashboard presents all agent sessions and can simultaneously show different levels of agent details. The upper panel shows a team of agents working on image analysis tasks, featuring shared file storage, execution wall-time, file preview, and activity feed. The lower panel shows a team of agents working on medical question-answering tasks, featuring terminal actions and API-call-level messages.}
  \label{fig:townview}
\end{figure*}

\section{Introduction}
\label{sec:intro}
Tool-using LLM agents have greatly advanced in recent years, often capable of running for hours or days on end.
These agents can autonomously tackle tasks that used to require full human attention, such as completing software engineering tasks
end-to-end~\citep{wang2025openhands,huang2024mlagentbench,
chan2025mlebench}, generating research hypotheses and running
experiments~\citep{luaiscientist,gottweisaicoscientist,kon2025curie,jiang2025aide}, and computer use~\citep{anthropic2024computeruse,openai2025operator}.

Greater capabilities bring about greater complexity.
A long-running agent leaves behind a messy transcript of interleaved reasoning steps, tool calls, and file accesses.
For a human, supervising the agent's actions can get non-trivial, and spending time studying agent traces partly defeats the time-savings promised by delegation. On the other hand, not spending the effort to understand agent actions hinders the opportunity to improve and customize agent behavior.

Therefore, we introduce \sys{}, an open-source, local GUI for observing and managing fleets of long-running AI agents.
It combines real-time trajectory visualization, manual and automated steering, and multi-agent coordination.
Trajectories of agents with different harnesses, such as open-source Hermes agent~\citep{hermesagent}, and proprietary Claude Code agent~\citep{anthropic2025claudecode}, are unified in the same lightweight interface.
We demonstrate how AgentGUI helps users understand complex agent trajectories $38\%$ faster, and increases proof-of-concept task completion rate of locally hosted agent by up to 34 percentage points.
Together, \sys{} provides a human-centered approach to keep agents easily manageable and supervisable in personal workflows.

\section{Related Work}
\label{sec:related}
\subsection{LLM Agents and Harness}
LLM agents are systems that use language models to reason, select tools, and take actions over multiple turns, often through interleaved reasoning traces~\citep{yao2023react} or executable code~\citep{wang2024codeact}.
The software environment enabling the agent to manage its context and act over long horizons is increasingly referred to as the harness. Notable frameworks include SWE-agent~\citep{yang2024sweagent}, OpenHands~\citep{wang2025openhands}, OpenClaw~\citep{openclaw2026}, and Hermes~\citep{hermesagent}.

\subsection{Observing Agent Trajectories}
Humans are not the only ones to struggle to read machine-centered agent transcripts.
Frontier LLMs, tasked with debugging agent behavior from raw traces, localize only a small fraction of errors on the TRAIL benchmark~\citep{deshpande2025trail}.
A first line of work improves the visualization and diagnosis of trajectories.
Examples include Agent-flow, which renders a live coding-agent session as a branching graph of tool calls and subagent activity~\citep{patole2026agentflow}, and IBM's Agent Trajectory Explorer~\citep{desmond2025trajectory}.
For debugging, AgentDiagnose extracts and visualizes trajectory statistics~\citep{ou2025agentdiagnose}, and AgentLens scales such analytics to multi-agent simulation histories~\citep{lu2024agentlens}.
These systems make agent behavior more legible, but provide no mechanism for intervening in or redirecting an ongoing run.

\subsection{Steering Agents at Runtime}
A second line of work contributes to steering agent behavior.
Operating on the AutoGen framework~\citep{wu2024autogen}, AutoGen Studio provides a no-code builder and debugger for multi-agent workflows~\citep{dibia2024autogenstudio}, and AGDebugger adds interactive message editing and resets for steering multi-agent teams~\citep{epperson2025agdebugger}.
Magentic-UI supports co-planning and co-tasking between human and agent~\citep{bansal2025magenticui}.
ResearStudio streams a deep-research agent's plan and actions to a live interface where the user can intervene~\citep{yang2025researstudio}.

Despite prior works, observability, steering, and collaboration between open-harness agents rarely co-occur.
Trajectory visualizers and debuggers often target explainability but leave out steering, whereas steering interfaces are often bound to a specific harness. 
Concurrently, Munder Difflin~\citep{giri2026munderdifflin} coordinates sessions of coding agents as avatars in a local office interface, rendering each agent's native terminal stream rather than a unified trajectory representation across harnesses.

\begin{figure*}[t]
  \centering
  \begin{tikzpicture}
    \node[anchor=south west,inner sep=0] (img)
      {\includegraphics[width=\textwidth]{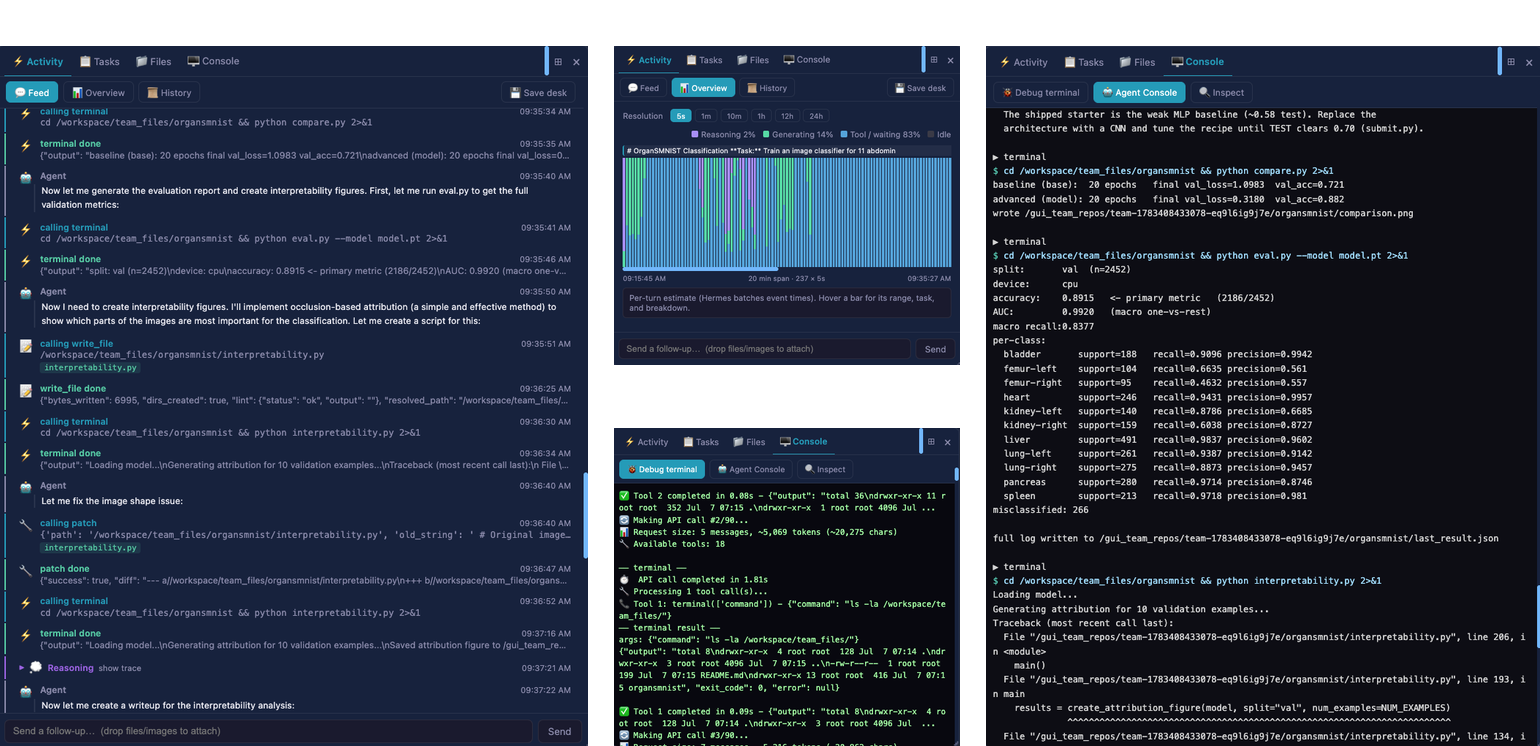}};
    \begin{scope}[x={(img.south east)},y={(img.north west)}]
      \node[anchor=base west,inner sep=0] at (0.0000,0.9491) {\panel{a}};
      \node[anchor=base west,inner sep=0] at (0.3987,0.9491) {\panel{b}};
      \node[anchor=base west,inner sep=0] at (0.3987,0.4370) {\panel{c}};
      \node[anchor=base west,inner sep=0] at (0.6403,0.9491) {\panel{d}};
    \end{scope}
  \end{tikzpicture}
  {\phantomsubcaption\label{fig:feat-feed}}
  {\phantomsubcaption\label{fig:feat-overview}}
  {\phantomsubcaption\label{fig:feat-debug}}
  {\phantomsubcaption\label{fig:feat-console}}
\caption{Per-desk views of a single run. \panel{a}~Activity feed: a live event timeline with visual separation between different types of actions. \panel{b}~Overview: a wall-clock timeline of the trajectory. \panel{c}~Debug terminal: per-call API and token telemetry. \panel{d}~Agent console: the terminal/code execution stream.}
  \label{fig:features}
\end{figure*}
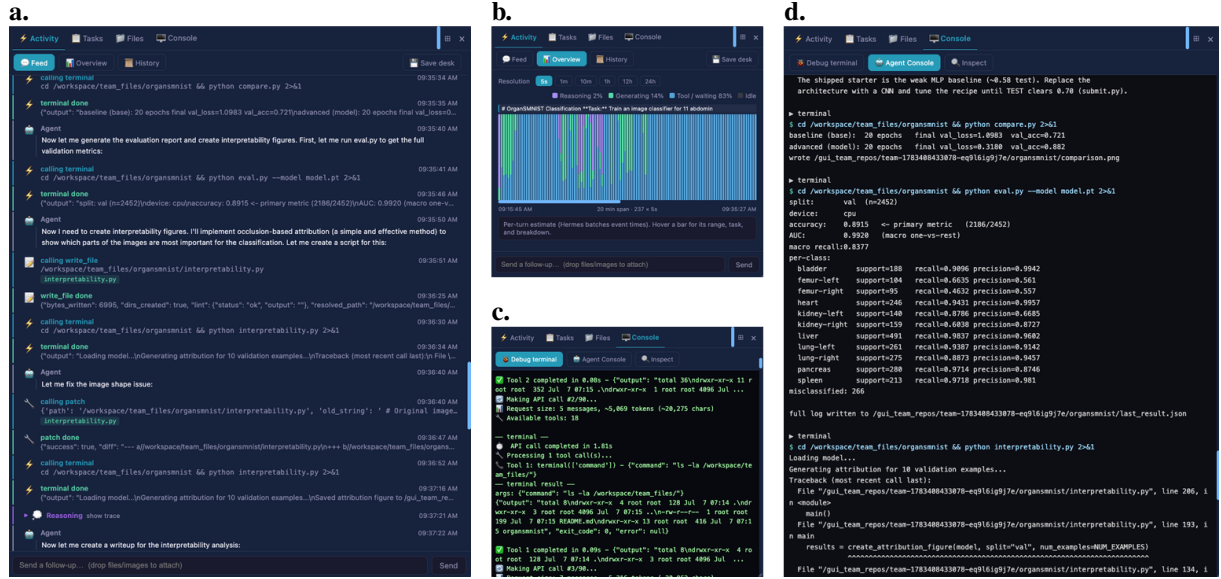

\section{\sys{}}
\label{sec:system}

\sys{} targets researchers and developers who run fleets of agents across open-source harnesses and need to monitor their behavior, detect and correct drift, and coordinate collaboration among them.
We describe \sys{} through a user’s journey, from managing agents on the dashboard to observing and steering individual agents, and then highlight several notable engineering features.

\subsection{Agent Configuration and Collaboration}
\label{sec:fleet}
The \sys{} dashboard (Fig.~\ref{fig:townview}) presents running agents as workers in a pixel-art office, with each agent assigned its own desk.
Starting a task is as simple as selecting an empty desk and entering a prompt, and optionally a single drag-and-drop action to provide context files.
Primary support targets Hermes agents, and experimental support covers the Claude Agent SDK~\citep{anthropic2025agentsdk}, which exposes the agent loop behind Claude Code.
Users can customize agent profiles, including their memory, system prompts, model configurations, and tool settings, directly from the GUI.
Agents working on the same task can be grouped into teams and collaborate through artifact sharing, enabling use cases e.g. a powerful agent refactoring the code base from a long session of a local agent.

\subsection{Agent Observation}
\label{sec:observation}
An agent work desk consists of 4 major tabs, visualizing agent activity, task definition, workspace files, and debug messages.
Agent trajectory is displayed in 4 minor tabs with varying levels of detail.
The activity feed (Fig.~\ref{fig:feat-feed}) decomposes agent traces into reasoning and generation content, and tool requests and responses, with distinct visual cues for skimming.
The overview feed (Fig.~\ref{fig:feat-overview}) renders the time agent spends on each task.
The console tab includes a comprehensive turn-level debug log with token telemetry (Fig.~\ref{fig:feat-debug}), as well as a lightweight, code-centered terminal console (Fig.~\ref{fig:feat-console}).
The latter helps heavy coding agent users to quickly comprehend important actions.
Sub-agents spawned by Hermes agent's delegate tool are visualized as mini expandable avatars next to the main agent's desk, with traces available.
Additionally, the files tab offers one-click previews of agent work directories and artifacts.

\subsection{Agent Steering}
\label{sec:steering}
\begin{figure*}[t]
  \centering
  \begin{tikzpicture}
    \node[anchor=south west,inner sep=0] (img)
      {\includegraphics[width=\textwidth]{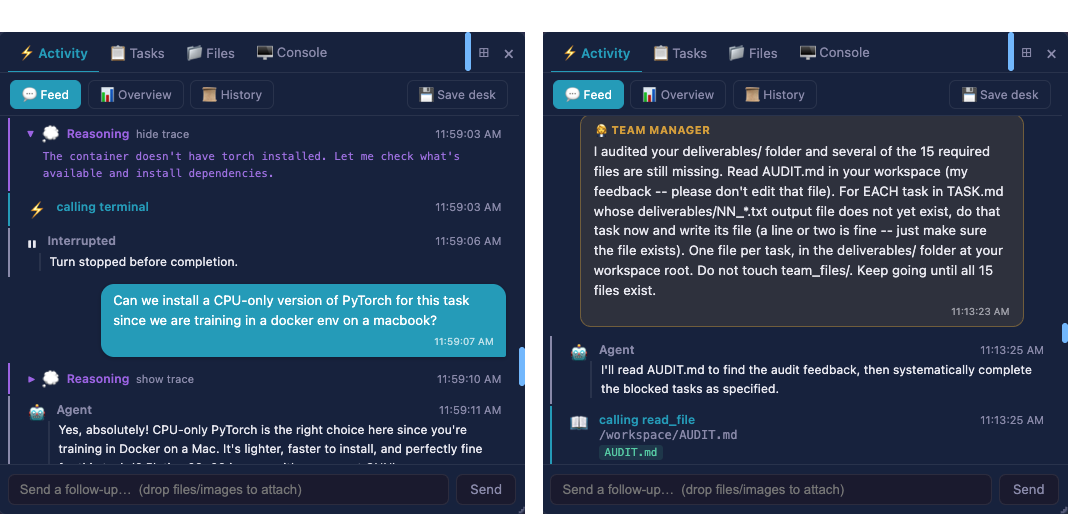}};
    \begin{scope}[x={(img.south east)},y={(img.north west)}]
      \node[anchor=base west,inner sep=0] at (0.0000,0.9494) {\panel{a}};
      \node[anchor=base west,inner sep=0] at (0.5084,0.9494) {\panel{b}};
    \end{scope}
  \end{tikzpicture}
  {\phantomsubcaption\label{fig:steer-user}}
  {\phantomsubcaption\label{fig:steer-manager}}
  \caption{
  Manual and automated steering channels in \sys{}. \panel{a}~User intervention: user message interrupts and redirects the agent's current turn. \panel{b}~Manager audit: a manager audit detects drift and auto-resumes the agent with corrective feedback.
  }
  \label{fig:steering}
\end{figure*}

A human can intervene and redirect the agent's current turn by direct input (Fig.~\ref{fig:steer-user}).
One can also modify the agent's task definition from the task tab, which will be reviewed by the agent after the current turn completes.
One implicit steering method includes switching the agent profile during the same task's execution, so a more powerful model could take over a stalled task, or a local model could take over a monitoring task.

An LLM-powered automated manager audits agent trajectory and artifacts to steer when necessary (Fig.~\ref{fig:steer-manager}), intervening upon a user-initiated audit, or on a configurable interval when a desk is idle and not yet marked solved.
The manager first decomposes the agent task into verifiable criteria, then gathers evidence from agent transcript and workspace files, and finally judges each criterion against the evidence and leaves an audit report.
A session is either marked solved, or prompted to resume by reading the manager's audit output.

\subsection{Engineering Highlights}
\label{sec:impl}
We took extensive engineering measures to ensure accessibility for users with varying setups, and security.
\sys{} provides quick-start instructions for running agents from entirely on the user's hardware using Ollama~\citep{ollama}, to using remote GPU servers and hosted inference options.
Inherited from Hermes implementation, each desk owns a persistent Docker sandbox that isolates agents from the host, and from one another.
A locally hosted FastAPI server executes each agent turn in an isolated worker process and streams events to a React frontend over WebSockets.
Claude Code agents can instead use the user's existing Claude subscription without requiring a separate API key and billing.
Desks can be saved and fully restored, including both trajectories and workspaces snapshots for portability and sharing.

\section{System Evaluation}
\label{sec:study}

\subsection{User Study: Trajectory Comprehension}
\begin{figure*}[t]
  \centering

  \captionsetup{singlelinecheck=false,justification=raggedright}
  \begin{subfigure}[t]{\textwidth}
    \centering
    \caption*{\panel{a}}
    \includegraphics[width=\textwidth]{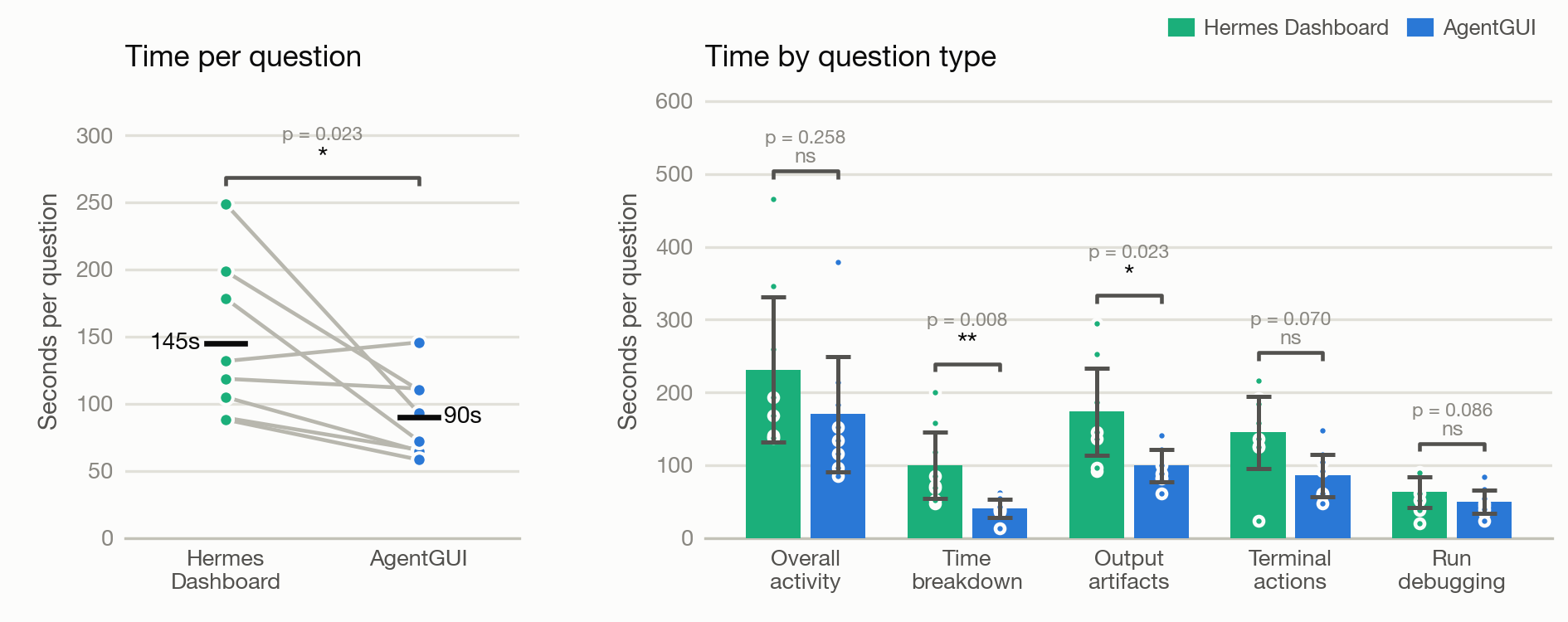}
    \phantomsubcaption\label{fig:study-time}
  \end{subfigure}
  \vspace{0.7em}
  \begin{subfigure}[t]{\textwidth}
    \centering
    \caption*{\panel{b}}
    \includegraphics[width=\textwidth]{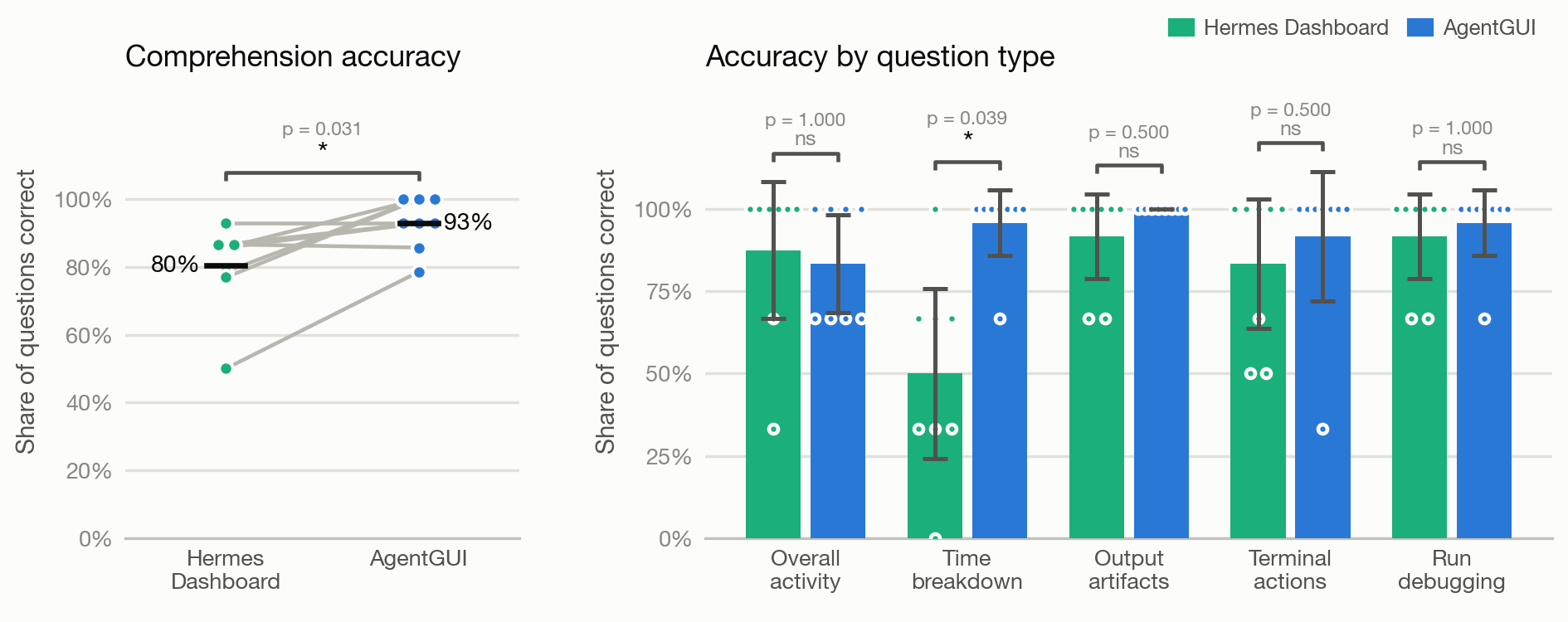}
    \phantomsubcaption\label{fig:study-acc}
  \end{subfigure}
  \vspace{0.7em}
  \begin{subfigure}[t]{\textwidth}
    \centering
    \caption*{\panel{c}}
    \includegraphics[width=\textwidth]{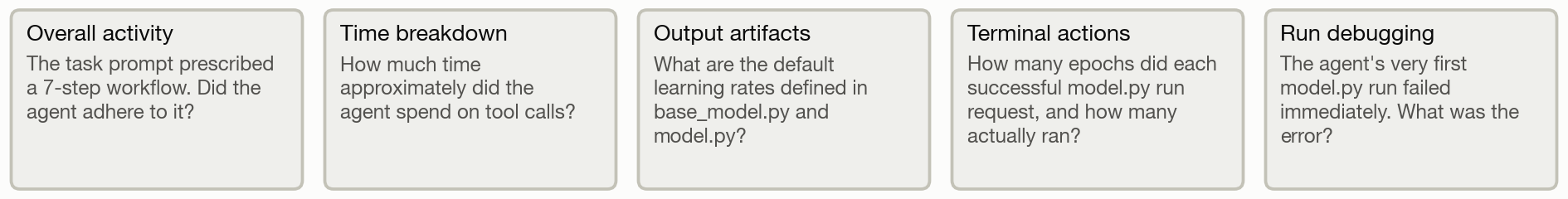}
    \phantomsubcaption\label{fig:study-questions}
  \end{subfigure}

  \captionsetup{singlelinecheck=true,justification=justified}
  \caption{
  User-study results ($N{=}8$, within-participant). \panel{a}~Time: mean seconds per question, per participant and interface (left; grey lines connect a participant's two sessions, black ticks are interface means) and by question type (right). \panel{b}~Accuracy: share of questions answered correctly, same layout. Brackets are exact paired sign-flip permutation tests on within-participant deltas; bars are means across sessions, error bars 95\% $t$-CIs across participants, dots individual sessions. \panel{c}~An example question for each type of question.}
  \label{fig:study}
\end{figure*}

\begin{figure}[t]
  \centering
  \includegraphics[width=\columnwidth]{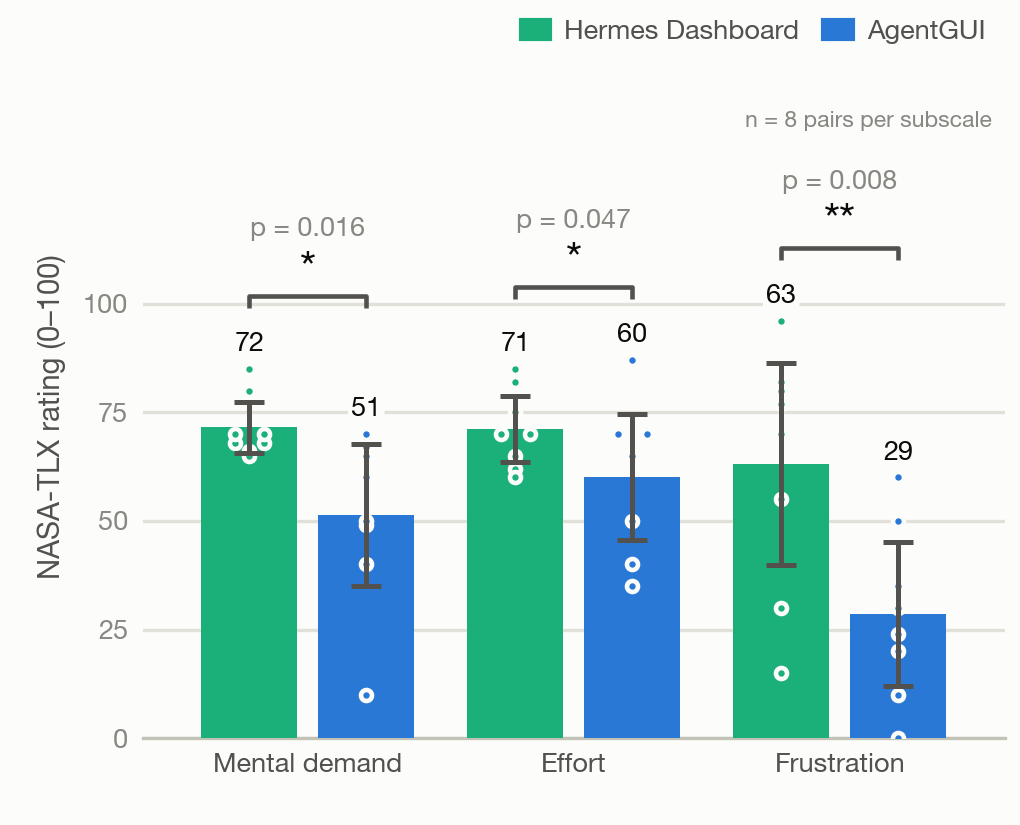}
  \caption{Self-reported workload under \sys{} and the Hermes Dashboard ($N{=}8$). Lower scores indicate lower workload; dots show participants, bars show means, and error bars show 95\% confidence intervals. Brackets report exact paired permutation-test $p$-values.
  }
  \label{fig:tlx}
\end{figure}

Does \sys{} help users better understand agent trajectories?
We measured the time and accuracy with which $N{=}8$ participants identified key information from agent trajectories. The baseline compared against is Hermes Dashboard (v0.16.0), a native visualization tool for Hermes Agent trace.

\paragraph{Design}
We defined two research tasks: training a CNN on OrganSMNIST~\citep{yang2023medmnist} against a frozen scorer, and iterating a system prompt for answering MedXpertQA~\citep{zuo2025medxpertqa} questions.
For each task, we generated two rollouts with a Hermes agent on a Qwen3.5-27B backbone~\citep{qwen35blog} (Appendix Table~\ref{tab:traces}), so that no comparison hinges on a single idiosyncratic trace.
For each rollout, we authored 14--15 questions, around 2--3 each on the agent's overall activity, time breakdown, output artifacts, terminal actions, and run debugging (example questions in Fig.~\ref{fig:study-questions}).
Each participant answered questions on two trajectories, one from each research task, one viewed in \sys{} and one in the dashboard, so that memorization could not carry over between interfaces.
Interface order and rollout assignment were counterbalanced: four participants saw \sys{} first, and each rollout was seen by exactly two participants per interface (assignment in Appendix Fig.~\ref{fig:study-design}).
To reduce noise from unfamiliarity with either visualizer, each participant was given five minutes of UI exploration, and the quiz included guidance on the location of relevant information.
Both interfaces exposed the same information categories (Appendix Table~\ref{tab:conditions}).
The study thus measures the interfaces' support for trajectory comprehension and information lookup, rather than familiarity with a particular UI.

\paragraph{Results}
Participants completed questions 38\% faster with \sys{} than with the baseline interface, taking on average 90 s rather than 145 s per question ($p = 0.023$, Appendix~\ref{sec:appendix-measures}, Fig.~\ref{fig:study}).
Completion time reduces across all five question types (Fig.~\ref{fig:study-time}), with statistical significance in time breakdown questions (59 s faster; $p = 0.008$) and output artifacts questions (74 s faster; $p = 0.023$).
Accuracy improved to 93\% from 80\% ($p = 0.031$, Fig.~\ref{fig:study}), although this was impacted by a low (50\%) score on the baseline interface.
Importantly, accuracy with \sys{} was not significantly worse in any question type (Fig.~\ref{fig:study}b).

We found no evidence of a speed–accuracy trade-off (left panels of Fig.~\ref{fig:study-time} and Fig.~\ref{fig:study-acc}).
Users report statistically significantly (Fig.~\ref{fig:tlx}) lower mental demand, frustration, and effort (adapted from the NASA-TLX~\citep{hart1988nasatlx}).

\subsection{Automated Steering against Drift}
\label{sec:manager}

\paragraph{Design}
We ran a proof-of-concept experiment to test whether the automated manager can improve task completion rate.
The task simulates an agent navigating through a synthetic patient chart consisting of 98 files, backed by a local open-source model to preserve data privacy.
The agent is asked to create 15 deliverables for 15 data aggregation tasks.
A programmatic scorer checks the presence of the deliverables, and a manager (Qwen3.5-27B) audits the workspace if the agent did not complete the task.
The workspace is scored again when the agent addresses the manager's comments.

\begin{figure}[t]
  \centering
  \includegraphics[width=\columnwidth]{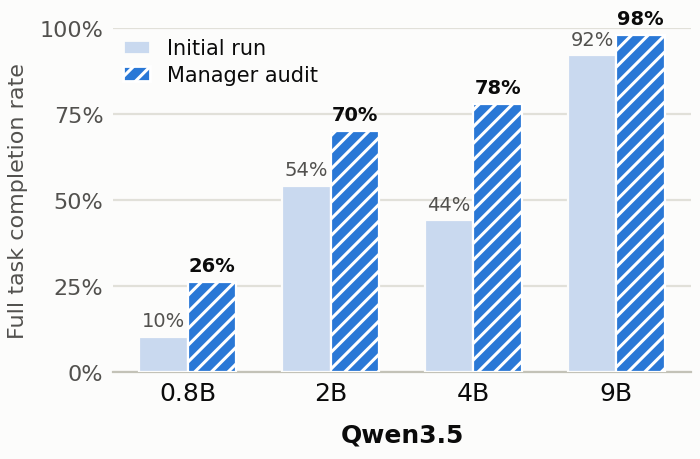}
  \caption{Effect of one manager audit on task completion across Qwen3.5 worker sizes ($N{=}50$ runs per model). Solid bars show completion before the audit, while hatched bars show completion after steering.
  }
  \label{fig:manager}
\end{figure}

\paragraph{Results}
Across $N{=}50$ runs per model size (Fig.~\ref{fig:manager}), initial task completion rates vary by model capability, but increases with model size.
The 4B's lower unaided completion likely reflects its tendency to use absolute path, which the sandbox write-guard rejects, so some of its deliverables fail to land until the manager audit flags the gap and prompts a corrected rewrite.
After a single audit, completion recovers a clean monotonic ordering in model size and improves at every scale—evidence that the benefit is general, not tied to any one worker.
The lift is largest where the worker leaves partial work, while saturating near weakest and strongest models: $10\%{\to}26\%$ (0.8B), $54\%{\to}70\%$ (2B), $44\%{\to}78\%$ (4B), and $92\%{\to}98\%$ (9B).

Overall, manager tokens are much cheaper than agent execution, making up no more than 1\% of the total token usage per model (Fig.~\ref{fig:token-budget}).
We observe additionally that the smallest local models spend more tokens than the larger, more capable ones on this task.
Larger models, e.g. the 9B model, has higher first-shot completion rate, invoking the manager not so frequently and uses fewer tokens.

\section{Conclusion}
\label{sec:conclusion}
We present \sys{}, an open-source (MIT License), locally hosted GUI for observing, steering, and coordinating fleets of long-running AI agents.
User study and proof-of-concept experiment highlights the effectiveness of observability, and to some extent, drift-prevention.
Observability is a precondition for trusting delegated agent work. An interface that makes agent activity legible and correctable can help deliver high quality outputs. We envision \sys{} to encourage future work on live supervision, broader harness support, and automated audits that target open-ended quality.

\section*{Limitations}
This study has several limitations.
First, the small and rather homogeneous user study limits the statistical test power and may not be fully generalizable. 
Second, the automated manager steering experiment focuses only on quantitative completion given small local model constraint.
Finally, we have only presented experiments to benchmark potential observability and automated steering improvement \sys{} offers. Experiments that require live human steering, preferably on a multi-agent scale and emphasize qualitative evaluation, would be an important future direction.

\section*{Ethics Statement}
Participation in the user study was voluntary. All participants gave informed consent through an in-app consent screen before starting the user study, and were free to withdraw at any time. Participants received no compensation; none are the authors of the manuscript.

\section*{Acknowledgments}
We thank the participants of our user study for their time and engagement. We thank Anthropic and the ETH AI Club for hosting the June 2026 Anthropic-ETH AI Sprint, during which part of the work was developed.

\bibliography{custom}

\clearpage

\appendix
\setlength{\intextsep}{6pt plus 2pt minus 2pt}
\setlength{\textfloatsep}{8pt plus 2pt minus 2pt}
\setlength{\floatsep}{6pt plus 2pt minus 2pt}
\renewcommand{\arraystretch}{0.94}

\section{User Study Setup}
\label{sec:appendix-design}

Each participant completed two quiz blocks.
Interface order and rollout assignment were counterbalanced: four of the eight participants started with \sys{}, and each rollout was read by exactly two participants per interface.

\begin{figure}[h]
  \centering
  \includegraphics[width=0.85\columnwidth,keepaspectratio]{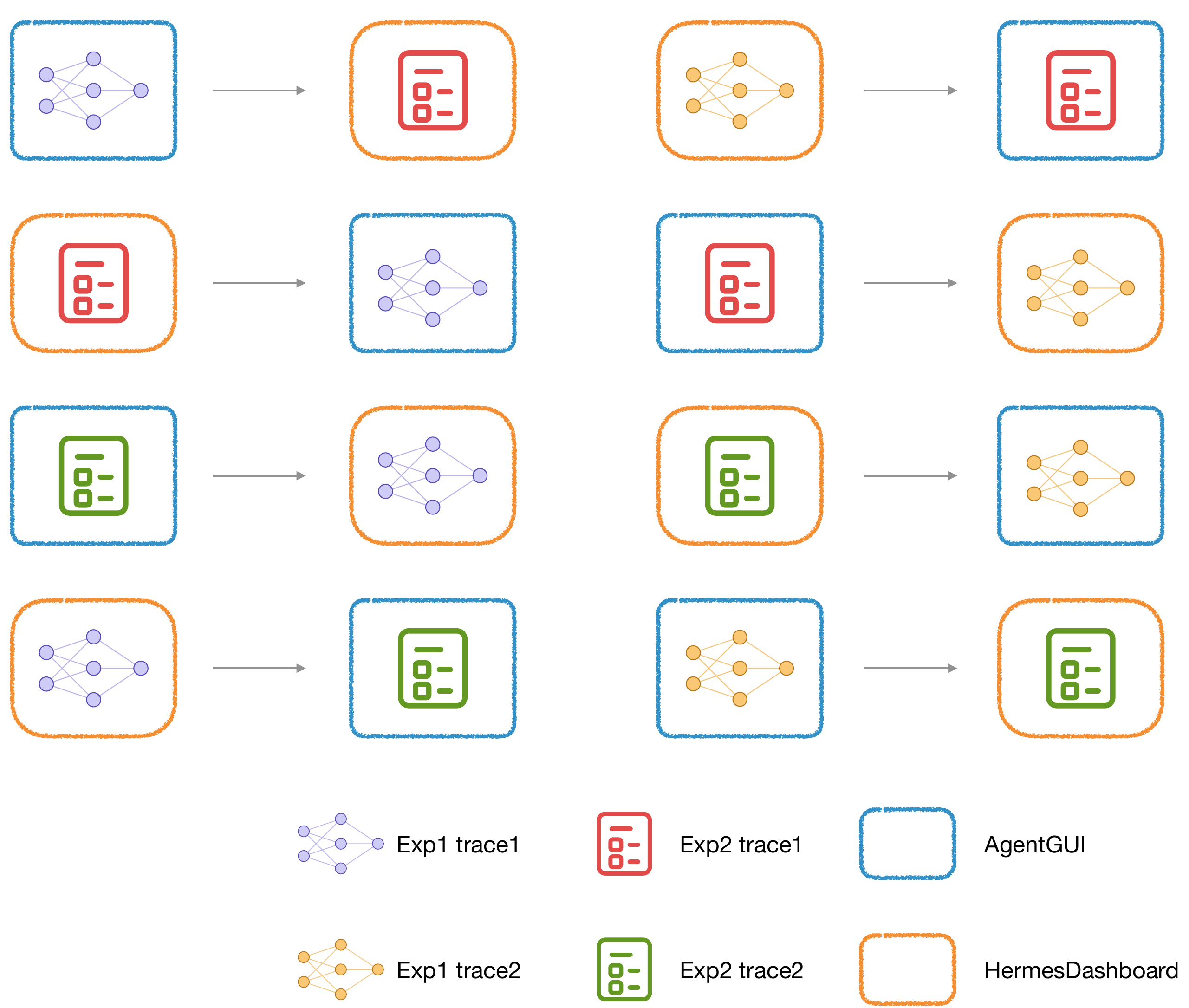}
  \caption{Participant--block assignment. Each pair is one participant; the two cards are their two quiz blocks, a model-training rollout (network icon) and a prompt-engineering rollout (clipboard icon), outlined in the colour of the interface they were read in, with the arrow pointing from the participant's first block to their second.}
  \label{fig:study-design}
\end{figure}

\section{Statistical Testing and Analysis}
\label{sec:appendix-measures}

Per rollout, we compute accuracy (the share of the rollout's questions answered correctly) and the seconds it took a participant to answer each question.
Each participant contributes one value per interface, and the unit of analysis is the within-participant difference $\Delta_i = \text{\sys{}}_i - \text{Dashboard}_i$, which cancels between-person variance in skill and reading speed.

Significance is assessed with the exact two-sided sign-flip permutation test on the mean delta: under the null hypothesis the interface labels are exchangeable within a participant, so all $2^{8}{=}256$ sign assignments of the observed deltas are equally likely, and
\[
p \;=\; \#\bigl\{\mathbf{s}\in\{\pm1\}^{8} : \bigl|\tfrac{1}{8}\textstyle\sum_i s_i\Delta_i\bigr| \ge |\bar{\Delta}|\bigr\} \,\big/\, 2^{8},
\]
i.e., the share of sign assignments whose mean is at least as extreme as the one observed.
Each contrast is reported with a 95\% paired-$t$ confidence interval on $\bar{\Delta}$.
\begin{table}[t]
\centering
\small
\setlength{\tabcolsep}{4.5pt}
\begin{tabular}{@{}lcc@{}}
\toprule
\textbf{Information} & \textbf{\sys{}} & \textbf{Hermes Dashboard} \\
\midrule
Message transcript       & Activity feed & Sessions tab \\
Tool calls + results     & Typed cards & Inline blocks \\
Time breakdown           & Overview chart & Message timestamps \\
Files written            & Files tab & In-app file directory \\
Terminal output          & Console tab & Inline blocks \\
\bottomrule
\end{tabular}
\caption{Trajectory information available in the two interfaces used in the user study. Both interfaces exposed the same five information categories but differed in how they organized and presented them.}
\label{tab:conditions}
\end{table}
One interrupted question was dropped (an 833\,s timer, due to the participant being interrupted by unforeseen circumstances); one question was excluded for all participants after a post-hoc review found it has no correct answer; two questions' accepted-answer sets were widened to two defensible readings; and one participant's first three questions are flagged for a hardware issue (kept in the primary analysis, excluded in a sensitivity variant).

\begin{table}[h]
\centering
\small
\setlength{\tabcolsep}{2.5pt}
\begin{tabular}{@{}llrp{3.55cm}@{}}
\toprule
\textbf{Task} & \textbf{Trace} & \textbf{Len} & \textbf{Behavioural Signature} \\
\midrule
Training & rollout 1 & 23\,min & \textsc{pass} (test 0.74); \code{pip} disk-full recovery; scored the sealed test twice \\
Training & rollout 2 & 20\,min & \textsc{pass} (0.72); uses a todo-list planner; 3 failed training runs\\
Prompt eng. & rollout 1 & 19\,min & \emph{overfit}: dev $1.00 \rightarrow$ test $0.42$; final prompt hardcodes dev answers, violating the prompt \\
Prompt eng. & rollout 2 & 40\,min & \emph{honest}: catches itself hardcoding mid-run and rewrites; dev $0.40$, test $0.35 \rightarrow 0.45$ \\
\bottomrule
\end{tabular}
\caption{Description of four agent rollouts used to generate study questions.}
\label{tab:traces}
\end{table}

\section{Cost Analysis For Manager Steering}
\label{sec:appendix-tokens}

\begin{figure}[!ht]
  \centering
  \includegraphics[width=0.85\columnwidth]{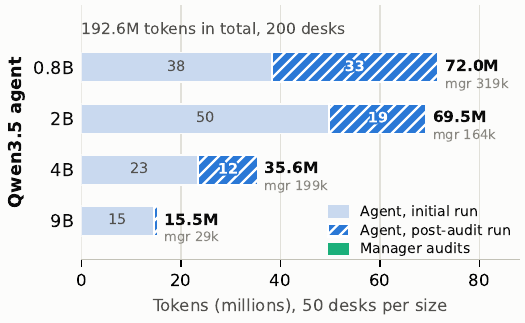}
  \caption{Token budget of the steering experiment, by agent size ($N{=}50$ desks each). Bars are stacked: the agent's initial unaided run, its run after the Manager's nudge, and the Manager's audit calls (0.19--0.56\% of each bar, printed at the bar end).}
  \label{fig:token-budget}
\end{figure}

Fig.~\ref{fig:token-budget} reports token costs of the experiment in Section~\ref{sec:manager}.
Agent tokens are the usage the serving endpoint reported for each completion, accumulated per desk.
Manager calls are tokenized using the tokenize endpoint on vllm.

\end{document}